%% file: main.tex
\documentclass[sigconf]{acmart}

\copyrightyear{2018} 
\acmYear{2018} 
\setcopyright{acmcopyright}
\acmConference[KDD '18]{The 24th ACM SIGKDD International Conference on Knowledge Discovery \& Data Mining}{August 19--23, 2018}{London, United Kingdom}
\acmBooktitle{KDD '18: The 24th ACM SIGKDD International Conference on Knowledge Discovery \& Data Mining, August 19--23, 2018, London, United Kingdom}
\acmPrice{15.00}
\acmDOI{10.1145/3219819.3220115}
\acmISBN{978-1-4503-5552-0/18/08}

\usepackage{subfigure, multirow, tabularx} 
\usepackage{booktabs} 
\usepackage{enumitem}
\usepackage{flushend}
\usepackage[T1]{fontenc}
\usepackage{epstopdf}
\usepackage{balance}
\usepackage{graphicx}
\usepackage[noend,ruled,linesnumbered]{algorithm2e} 
\usepackage{microtype}
\usepackage{url}
\usepackage{wrapfig,lipsum}
\usepackage{makecell}
\usepackage{wrapfig}

\newtheorem{thm:def}{Definition}
\newtheorem{thm:eg}{Example}
\newtheorem{thm:lem}{Lemma}
\newtheorem{thm:obs}{Observation}
\newtheorem{thm:req}{Requirement}

\newcommand{\mquote}[1]{{``\emph{#1}''}}

\DeclareMathAlphabet{\mathbbold}{U}{bbold}{m}{n}

\newcommand\defeq{\mathrel{\overset{\makebox[0pt]{\mbox{\normalfont\tiny\sffamily def}}}{=}}}

\DeclareMathOperator*{\argmax}{arg\,max}

\DeclareMathOperator{\ancestorF1}{\mathnormal{Ancestor}-\mathnormal{F1}}
\DeclareMathOperator{\edgeF1}{\mathnormal{Edge}-\mathnormal{F1}}

\input{macro.tex}

\begin{document}

\fancyhead{}

\title{HiExpan: Task-Guided Taxonomy Construction by \\ Hierarchical Tree Expansion}

\author{Jiaming Shen$^{1}$, Zeqiu Wu$^{1}$\footnotemark[1], Dongming Lei$^{1}$\footnotemark[1], Chao Zhang$^1$, Xiang Ren$^2$, \\ Michelle T. Vanni$^{3}$, Brian M. Sadler$^{3}$, Jiawei Han$^1$
}
\authornote{These two authors have equal contributions.}
\affiliation{%
  \institution{$^1$Department of Computer Science, University of Illinois at Urbana-Champaign, IL, USA}
}
\affiliation{%
  \institution{$^2$Department of Computer Science, University of Southern California, CA, USA}
}
\affiliation{%
  \institution{$^3$U.S. Army Research Laboratory, MD, USA}
}
\affiliation{%
  \institution{$^1$\{js2, zeqiuwu1, dlei5, czhang82, hanj\}@illinois.edu $\quad$ $^2$xiangren@usc.edu $\quad$}
}
\affiliation{%
  \institution{$^3$\{michelle.t.vanni.civ, brian.m.sadler6.civ\}@mail.mil}
}

\begin{abstract}
    \input{0-abstract.tex}

\end{abstract}


\maketitle

\input{1-introduction.tex}
\input{2-related_work.tex}

\input{3-problem.tex}
\input{4-methodology.tex}

\input{5-experiments.tex}
\input{6-conclusion.tex}
\input{7-ack.tex}

\bibliographystyle{ACM-Reference-Format}
\bibliography{cited}

\end{document}

%% file: macro.tex
\usepackage{amsmath, amsfonts, amssymb, amsthm, xspace, color}

\newcommand{\hide}[1]{} 

\newcommand{\nop}[1]{}

\newcommand{\etal}{\emph{et~al.}\xspace} 
\newcommand{\ie}{\emph{i.e.}\xspace} 
\newcommand{\eg}{\emph{e.g.}\xspace} 

\def \v {\mathbf{v}}

\def \C {\mathcal{C}}
\def \D {\mathcal{D}}
\def \E {\mathcal{E}}

\def \R {\mathcal{R}}

\def \T {\mathcal{T}}

\newcommand{\AutoPhrase}{\mbox{\sf AutoPhrase}\xspace}

\newcommand{\HiExpan}{\mbox{\sf HiExpan}\xspace}
\newcommand{\SetExpan}{\mbox{\sf SetExpan}\xspace}

\newcommand{\widthExpan}{\textsc{Width-Expansion}\xspace}
\newcommand{\depthExpan}{\textsc{Depth-Expansion}\xspace}
\newcommand{\HSetExpan}{\mbox{\sf HSetExpan}\xspace}
\newcommand{\NoREPEL}{\mbox{\sf NoREPEL}\xspace}
\newcommand{\NoGTO}{\mbox{\sf NoGTO}\xspace}
\newcommand{\Sg}{\mbox{skip-pattern}\xspace}
\newcommand{\Sgs}{\mbox{skip-patterns}\xspace}

%% file: 0-abstract.tex

Taxonomies are of great value to many knowledge-rich applications.  
As the manual taxonomy curation costs enormous human effects, automatic taxonomy construction is in great demand. 
However, most existing automatic taxonomy construction methods can only build hypernymy taxonomies wherein each edge is limited to expressing the ``is-a'' relation.  
Such a restriction limits their applicability to more diverse real-world tasks where the parent-child may carry different relations. 
In this paper, we aim to construct a \emph{task-guided taxonomy} from a domain-specific corpus, and allow users to input a ``seed'' taxonomy, serving as the task guidance.  
We propose an expansion-based taxonomy construction framework, namely \HiExpan, which automatically generates key term list from the corpus and iteratively grows the seed taxonomy.  
Specifically, \HiExpan views all children under each taxonomy node forming a coherent set and builds the taxonomy by recursively expanding all these sets.  
Furthermore, \HiExpan incorporates a weakly-supervised relation extraction module to extract the initial children of a newly-expanded node and adjusts the taxonomy tree by optimizing its global structure.  
Our experiments on three real datasets from different domains demonstrate the effectiveness of \HiExpan for building task-guided taxonomies.

%% file: 1-introduction.tex
\section{Introduction}\label{sec:intro}

\begin{figure}[!t]
  \centering
  \centerline{\includegraphics[width=0.48\textwidth]{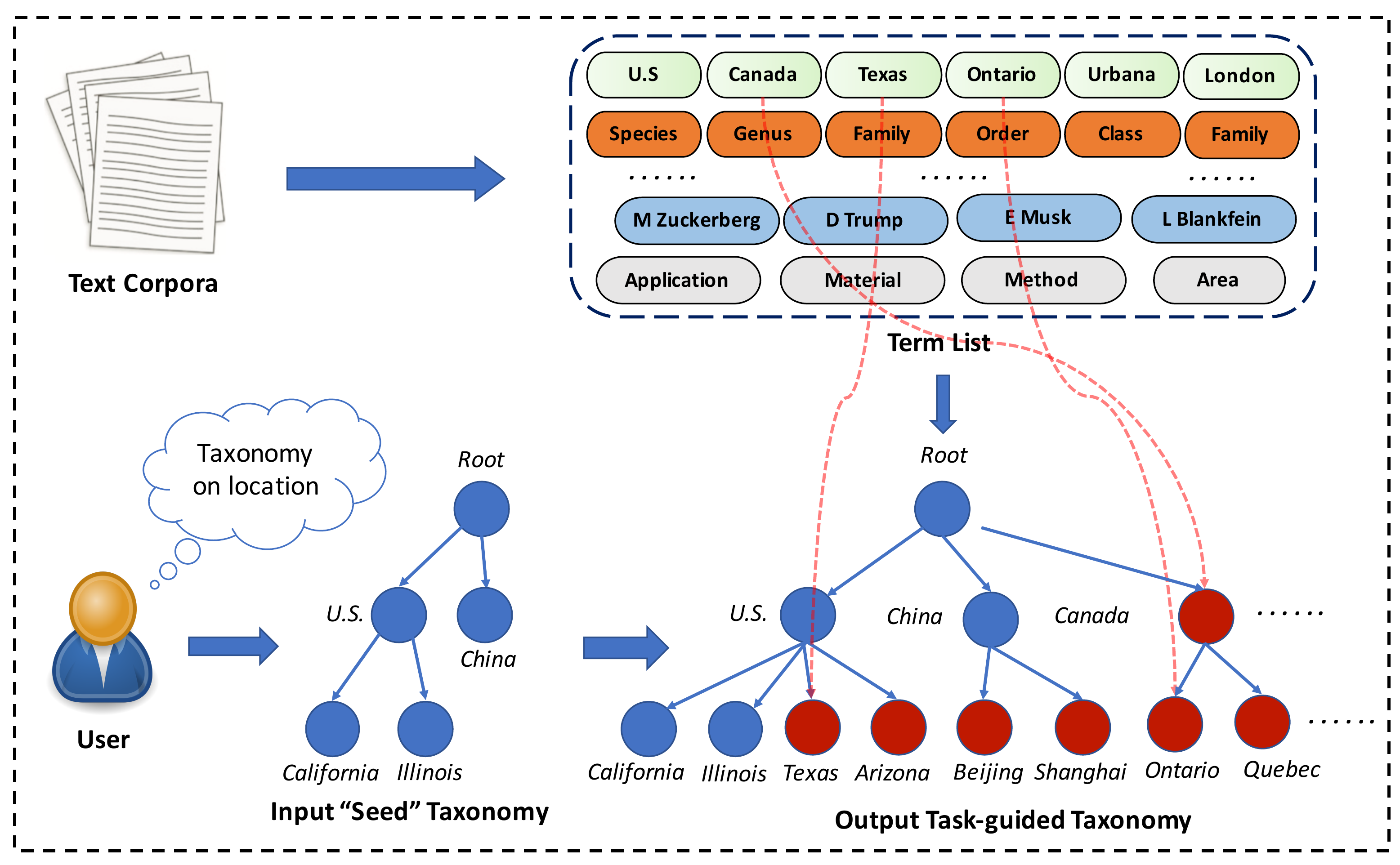}}
  \vspace{-0.1cm}
  \caption{Task-guided taxonomy construction. User provides a ``seed'' taxonomy tree as task guidance, and we will extract key terms from raw text corpus and generates the desired taxonomy automatically.}
  \label{fig:motivation}
\end{figure}

Taxonomy is the backbone of many knowledge-rich applications such as question answering \cite{Yang2017EfficientlyAT}, query understanding \cite{Hua2017UnderstandST}, and personalized recommendation \cite{Zhang2014TaxonomyDF}.
Most existing taxonomies are constructed by human experts or in a crowd-sourcing manner.
However, such manual constructions are labor-intensive, time-consuming, unadaptable to changes, and rarely complete.
As a result, automated taxonomy construction is in great demand.

Existing methods mostly build taxonomies based on ``is-A'' relations (\eg, a \mquote{panda} is a \mquote{mammal} and a \mquote{manmal} is an \mquote{animal}) \cite{Wu2012ProbaseAP, Velardi2013OntoLearnRA, Wang2017ASS} by first leveraging pattern-based or distributional methods to extract hypernym-hyponym term pairs and then organizing them into a tree-structured hierarchy.
However, such hierarchies cannot satisfy many real-world needs due to its 
(1) \emph{inflexible semantics}: many applications may need hierarchies carrying more flexible semantics such as ``\emph{city-state-country}" in a location taxonomy; and
(2) \emph{limited applicability:} the ``universal" taxonomy so constructed is unlikely to fit diverse and user-specific application tasks.  

This motivates us to work on \emph{task-guided} taxonomy construction, which takes a user-provided ``seed'' taxonomy tree (as task guidance) along with a domain-specific corpus and generates a desired taxonomy automatically.  
For example, a user may provide a seed taxonomy containing only two countries and two states along with a large corpus, and our method will output a taxonomy which covers all the countries and states mentioned in the corpus.

In this study, we propose \HiExpan, a framework for task-guided taxonomy construction. 
Starting with a tiny seed taxonomy tree provided by a user, a weakly supervised approach can be developed by set expansion.
A set-expansion algorithm aims to expand a small set of seed entities into a complete set of entities that belong to the same semantic class \cite{Rong2016egoset,Shen2017SetExpanCS}.
Recently we developed an interesting \SetExpan algorithm \cite{Shen2017SetExpanCS}, which expands a tiny seed set (\eg, \{\mquote{Illinois}, \mquote{California}\}) into a complete set (\eg, U.S. states mentioned in the corpus) by a novel bootstrapping approach.
While such an approach is intuitive, there are two major challenges by extending it to generating high-quality taxonomy:
(1) modeling global taxonomy information: a term that appears in multiple expanded sets may need conflict resolution and hierarchy adjustment accordingly, and
(2) cold-start with empty initial seed set: as an example, initial seed set \{\mquote{Ontario}, \mquote{Quebec}\} will need to be found once we add \mquote{Canada} at the country level as shown in Figure \ref{fig:motivation}.

\HiExpan consists of two novel modules for dealing with the above two challenges.
First, whenever we observe a conflict (\ie, the same term appearing in multiple positions on taxonomy) during the tree expansion process, we measure a ``confidence score" for putting the term in each position and select the most confident position for it. 
Furthermore, at the end of our hierarchical tree expansion process, we will do a global optimization of the whole tree structure.
Second, we incorporate a weakly-supervised relation extraction method to infer parent-child relation information and to find seed children terms under a specific parent.
Equipped with these two modules, \HiExpan constructs the task-guided taxonomy by iteratively growing the initial seed taxonomy tree.
At each iteration, it views all children under a non-leaf taxonomy node as a coherent set and builds the taxonomy by recursively expanding these sets.
Whenever a node with no initial children nodes found, it will first conduct seeds hunting.
At the end of each iteration, \HiExpan detects all the conflicts and resolves them based on their confidence scores.

In summary, this study makes the following contributions: 
\begin{enumerate}[leftmargin=*] 
  \item We introduce a new research problem \emph{task-guided taxonomy construction}, which takes a user-provided seed taxonomy along with a domain-specific corpus as input and aims to output a desired taxonomy that satisfies user-specific application tasks. 
  \item We propose \HiExpan, a novel expansion-based framework for task-guided taxonomy construction. \HiExpan generates the taxonomy by growing the seed taxonomy iteratively. Special mechanisms are also taken by \HiExpan to leverage global tree structure information.
  \item We conduct extensive experiments to verify the effectiveness of \HiExpan on three real-world datasets from different domains.
\end{enumerate}

The remaining of the paper is organized as follows.
Section~\ref{sec:related} discusses the related work.
Section~\ref{sec:problem} defines our problem.
Then, we present the \HiExpan framework in Section~\ref{sec:methodology}.
In Section~\ref{sec:exp}, we report and analyze the experimental results. 
Finally, we conclude the paper and discuss some future directions in Section~\ref{sec:con}.

%% file: 2-related_work.tex

\section{Related Work}\label{sec:related}

In this section, we review related work in following three categories. 

\subsection{Taxonomy Construction}
Most existing approaches to taxonomy construction focus on building hypernym-hyponym taxonomies wherein each parent-child pair expresses the ``is-a'' relation. 
Typically, they consist of two key steps: (1) hypernymy relation acquisition (\ie, obtaining hypernym-hyponym pairs), and (2) structured taxonomy induction (\ie, organizing all hypernymy relations into a tree structure).

Methods for hypernymy relation acquisition fall into two classes: pattern-based and distributional.  
One pioneering pattern-based method is Hearst patterns \cite{Hearst1992AutomaticAO} in which lexical syntactic patterns (\eg, ``$NP_{x}$ such as $NP_{y}$'') are leveraged to match hypernymy relations.
Later studies extend this method by incorporating more linguistic rules \cite{Snow2004LearningSP, Ritter2009WhatIT, Luu2014TaxonomyCU} or designing generalized patterns such as \mquote{star-pattern} \cite{Navigli2010LearningWL}, \mquote{SOL pattern} \cite{Nakashole2012PATTYAT}, and \mquote{meta-pattern} \cite{Jiang2017MetaPADMP}. 
These methods could achieve high precision in the result pairs but often suffer low recalls (\ie, many hypernym-hyponym pairs do not match the pre-defined patterns). 
Along another line, distributional methods predict whether a pair of terms $\langle x,y \rangle$ holds a hypernymy relation based on their distributional representations. 
Early studies first extract statistical features (\eg, the context words of a term), calculate pairwise term similarity using symmetric metrics (\eg, cosine, Jaccard) \cite{Lin1998AnID} or asymmetric metrics (\eg, WeedsPrec \cite{Weeds2004CharacterisingMO}, SLQS \cite{Roller2014InclusiveYS}), and predict if $\langle x,y \rangle$ holds a hypernymy relation.  
More recently, a collections of \emph{supervised} methods \cite{Baroni2012EntailmentAT, Fu2014LearningSH, Weeds2014LearningTD, Yu2015LearningTE, Luu2016LearningTE, Anke2016SupervisedDH} are proposed to
leverage pre-trained word embeddings and curated training data to directly learn a relation classification/prediction model.  
However, neither pattern-based nor distributional techniques can be applied to our problem because they are designed exclusively for acquiring hypernym-hyponym pairs, whereas we aim to construct a \emph{task-guided} taxonomy where the parent-child relations are task-specific and subject to user guidance.

For the structured taxonomy induction step, most methods first build a graph where edges represent noisy hypernymy relations, extracted in the former step, and then derive a tree-like taxonomy from this graph. 
Kozareva and Hovy \cite{Kozareva2010ASM} iteratively retain the longest paths between root and leaf terms and remove other conflicting edges.  
Navigli \etal~\cite{Navigli2011AGA} and Velardi \etal~\cite{Velardi2013OntoLearnRA} use the same longest-path idea to weigh edges and then find the largest-weight taxonomy as a Maximum Spanning Tree. 
Bansal \etal~\cite{Bansal2014StructuredLF} build a factor graph to model hypernymy relations and regard taxonomy induction as a structured learning problem, which can be inferred with loop belief propagation. 
Recently, Gupta \etal~\cite{Gupta2017TaxonomyIU} propose to build the initial graph using hypernym subsequence (instead of single hypernym pair) and model taxonomy induction as a minimum-cost flow problem \cite{Orlin1996APT}.  
Comparing with these methods, our approach leverages the weak supervision in ``seed'' taxonomy and builds a task-specific taxonomy in which two terms can hold a non-hypernymy relation. 
Further, our taxonomy construction framework jointly acquires task-specific relations and induces taxonomy structure, instead of performing the two tasks separately.

\subsection{Set Expansion}

Our work is also closely related to set expansion --- the task of expanding a small set of seed entities into a complete set of entities that belong to the same semantic class \cite{Wang2007LanguageIndependentSE}. 
One line of works, including \emph{Google Set} \cite{tong2008system}, \emph{SEAL} \cite{Wang2008IterativeSE} and Lyretail \cite{chen2016long}, solves this task by submitting a query of seed entities to an online search engine and mining top-ranked webpages. 
Other works aim to tackle the task in a \emph{corpus-based} setting where the set is expanded by offline processing a given corpus. 
They either perform a one-time ranking of all candidate entities \cite{Pantel2009WebScaleDS, Shi2010CorpusbasedSC, He2011SEISASE} or do iterative pattern-based bootstrapping \cite{Shi2014APC, Rong2016egoset, Shen2017SetExpanCS}. 
In this work, in addition to just adding new entities into the set, we go beyond one step and aim to organize those expanded entities in a tree-structured hierarchy (\ie, a taxonomy).

\subsection{Weakly-supervised Relation Extraction}
There have been studies on weakly supervised relation extraction, which aims at extracting a set of relation instances containing certain semantic relationships. 
Our method is related to corpus-level relation extraction that identifies relation instances from the entire text corpora \cite{Qu2018WeaklysupervisedRE,Zeng2015DistantSF,Mintz2009DistantSF,Riedel2013RelationEW}.
In the weakly supervised setting, there are generally two approaches for corpus-level relation extraction. 
The first is pattern-based \cite{Agichtein2000SnowballER, Jiang2017MetaPADMP, Nakashole2012PATTYAT}, which usually uses bootstrapping to iteratively extract textual patterns and new relation instances. 
The second approach \cite{Mikolov2013DistributedRO, Pennington2014GloveGV, Tang2015LINELI} tries to learn low-dimensional representations of entities such that entities with similar semantic meanings have similar representations.
Unfortunately, all these existing methods require a considerable amount of relation instances to train an effective relation classifier, which is infeasible in our setting as we only have a limited number seeds specified by users.
Furthermore, these studies do not consider organizing the relation pairs into a taxonomy structure.

%% file: 3-problem.tex

\section{Problem Formulation}\label{sec:problem}

The input for our taxonomy construction framework includes two parts: (1) a corpus $\D$ of documents; and (2) a ``seed'' taxonomy $\T^{0}$. 
The ``seed'' taxonomy $\T^{0}$, given by a user, is a tree-structured hierarchy and serves as the task guidance.  
Given the corpus $\D$, we aim to expand this seed taxonomy $\T^{0}$ into a more complete taxonomy $\T$ for the task. 
Each node $e \in \T$ represents a term\footnote{In this work, we use the word \mquote{term} and \mquote{entity} interchangeably.} extracted from corpus $\D$ and each edge $\langle e_{1}, e_{2} \rangle$ denotes a pair of terms that satisfies the task-specific relation.  
We use $\E$ and $\R$ to denote all the nodes and edges in $\T$ and thus $\T \defeq (\E, \R)$.

\begin{example}
  Figure \ref{fig:motivation} shows an example of our problem.  Given a
  collection of Wikipedia articles (\ie, $\D$) and a ``seed'' taxonomy
  containing two countries and two states in the \mquote{U.S.} (\ie,
  $\T^{0}=(\E^{0}, \R^{0})$), we aim to output a taxonomy $\T$ which covers all
  countries and states mentioned in corpus $\D$ and connects them based on the
  task-specific relation \mquote{located in}, indicated by $\R^{0}$.
\end{example}

%% file: 4-methodology.tex
\section{The HiExpan Framework}\label{sec:methodology}

In this section, we first give an overview of our proposed \HiExpan framework in Section \ref{subsec:framework_overview}. 
Then, we discuss our key term extraction module and hierarchical tree expansion algorithm in Section \ref{subsec:key_term_extraction} and Section \ref{subsec:hierarchical_tree_expansion}, respectively. 
Finally, we present our taxonomy global optimization algorithm in Section \ref{subsec:taxonomy_structure_opt}.

\subsection{Framework Overview}\label{subsec:framework_overview}

In short, \HiExpan views all children under each taxonomy node forming a coherent \emph{set}, and builds the taxonomy by recursively expanding all these sets.
As shown in Figure \ref{fig:motivation}, two first-level nodes (\ie, \mquote{U.S.} and \mquote{China}) form a set representing the semantic class \mquote{Country} and by expanding it, we can obtain all the other countries.
Similarly, we can expand the set \{\mquote{California}, \mquote{Illinois}\} to find all the other states in the U.S.

\begin{figure*}[!t]
  \centering
  \centerline{\includegraphics[width=1.0\textwidth]{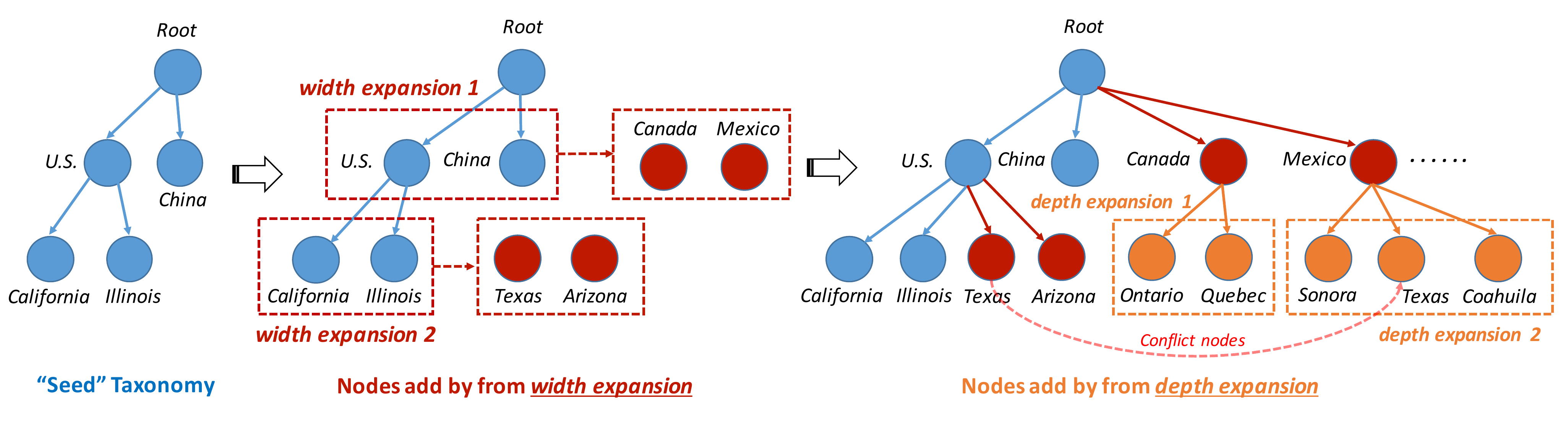}}
  \vspace{-0.3cm}
  \caption{An overview of our hierarchical tree expansion algorithm.}
  \label{fig:overview}
\end{figure*}

Given a corpus $\D$, we first extract all key terms using a phrase mining tool followed by part-of-speech filter.
Since the generated term list contains many task-irrelevant terms (\eg, people's names are totally irrelevant to a location taxonomy), we use a set expansion technique to carefully select best terms, instead of exhaustively testing all possible terms in the list.
We refer this process as \emph{width expansion} as it increases the \emph{width} of taxonomy tree.
Furthermore, to address the challenge that some nodes do not have an initial child (\eg, the node \mquote{Mexico} in Figure \ref{fig:overview}), we find the ``seed'' children by applying a weakly-supervised relation extraction method, which we refer as \emph{depth expansion}.
By iteratively applying these two expansion modules, our hierarchical tree expansion algorithm will first grow the taxonomy to its full size.
Finally, we adjust the taxonomy tree by optimizing its global structure.
In the following, we describe each module of \HiExpan in details.

\subsection{Key Term Extraction}\label{subsec:key_term_extraction}

We use \AutoPhrase, a state-of-the-art phrase mining algorithm \cite{Liu2015MiningQP, Shang2018AutomatedPM}, to extract all key terms in the given corpus $\D$.
\AutoPhrase outputs a key term list and identifies the in-corpus occurrences of each key term.
After that, we apply a Part-of-Speech (POS) tagger to the corpus and obtain the POS tag sequence of each key term occurrence.
Then, we retain the key term occurrence whose corresponding POS tag sequence contains a noun POS tag (\eg, \mquote{NN}, \mquote{NNS}, \mquote{NNP}).
Finally, we aggregate the key terms that have at least one remaining occurrence in the corpus into the key term list. 
Although the key term list so generated is noisy and may contain some task-irrelevant terms, recall is more critical for this step because we can recognize and simply ignore the false positives at the later stages of \HiExpan, but have no chance to remedy the mistakenly excluded task-relevant terms.

\subsection{Hierarchical Tree Expansion}\label{subsec:hierarchical_tree_expansion}

The hierarchical tree expansion algorithm in \HiExpan is designed to first grow the taxonomy tree.
It is based on (1) algorithm \SetExpan \cite{Shen2017SetExpanCS} which expands a small set of seed entities into a complete set of entities that belong to the same semantic class, and (2) REPEL \cite{Qu2018WeaklysupervisedRE} which utilizes a few relation instances (\ie, a pair of entities satisfying a target relation) as seeds to extract more instances of the same relation.
Our choice of these two algorithms is motivated by their effectiveness to leverage the weak supervision in the tiny ``seed'' taxonomy $\T^{0}$ specified by a user.

\subsubsection{Width Expansion}\label{subsubsec:width_expansion}
Width expansion aims to find the sibling nodes of a given set of children nodes which share the same parent, as demonstrated in the following example.

\begin{example}[Width Expansion]
Figure \ref{fig:overview} shows two expected width expansion results.
When given the set \{\mquote{U.S.}, \mquote{China}\}, we want to find their sibling nodes, \mquote{Canada}, \mquote{Mexico}, and put them under parent node \mquote{Root}.
Similarly, we aim to find all siblings of \{\mquote{California}, \mquote{Illinois}\} and attach them under parent node \mquote{U.S.}.
\end{example}

This naturally forms a set expansion problem and thus we adapt the \SetExpan algorithm in \cite{Shen2017SetExpanCS} for addressing it.
Compared with original \SetExpan algorithm, the width expansion algorithm in this paper incorporates the term embedding feature and better leverages the entity type feature.
In the following, we first discuss different types of features and similarity measures used, and then describe the width expansion algorithm in details.

\smallskip
\noindent \textbf{Features.} We use the following three types of features:

\begin{itemize}[leftmargin=*] 
  \item \emph{\Sg}\footnote{This feature was originally referred as ``skip-gram" feature in \cite{Shen2017SetExpanCS}. Here we change the terminology to avoid the confusion with the SkipGram model used in word2vec~\cite{Mikolov2013DistributedRO} for training word embeddings.}: Given a target term $e_{i}$ in a sentence, one of its \Sg features is ``$w_{-1}$ $\underline{\hspace{0.1in}}$ $w_{1}$'' where $w_{-1}$ and $w_{1}$ are two context words and $e_{i}$ is replaced with a placeholder.
  One advantage of \Sg feature is that it imposes strong positional constraints.
For example, one \Sg of term \mquote{California} in sentence \mquote{We need to pay California tax.} is \mquote{pay $\underline{\hspace{0.1in}}$ tax}.
Following \cite{Rong2016egoset, Shen2017SetExpanCS}, we extract up to six \Sgs of different lengths for one target term $e_{i}$ in each sentence.

  \item \emph{term embedding}: We use either the SkipGram model in word2vec \cite{Mikolov2013DistributedRO} or REPEL \cite{Qu2018WeaklysupervisedRE} (described in Section \ref{subsubsec:depth_expansion}) to learn the term embeddings.
We will first use \mquote{$\_$} to concatenate tokens in a multi-gram term (\eg, \mquote{Baja California}) and then learn the embedding of this term.
The advantage of term embedding feature is that it captures the semantics of each term.

  \item \emph{entity type}: We obtain each entity's type information by linking it to Probase \cite{Wu2012ProbaseAP}. The return types serve as the features of that entity. For entities that are not linkable, they simply do not have this entity type feature. 
\end{itemize}

\smallskip
\noindent \textbf{Similarity Measures.}
A key component in width expansion algorithm is to compute the sibling similarity of two entities $e_1$ and $e_2$, denoted as $sim_{sib}(e_1, e_2)$.
We first assign the weight between each pair of entity and \underline{sk}ip-pattern as follows:
\begin{equation}\label{eq:skip-pattern-weight}
\small
f_{e, sk} = \log(1+X_{e, sk}) \left[  \log|V| - \log(\sum_{e'} X_{e', sk}) \right],
\end{equation}
where $X_{e, sk}$ is the raw co-occurrence count between entity $e$ and \Sg $sk$, and $|V|$ is the total number of candidate entities.

Similarly, we can define the association weight between an entity and a \underline{ty}pe as follows:
\begin{equation}\label{eq:type-weight}
\small
f_{e, ty} = \log(1+C_{e, ty}) \left[  \log|V| - \log(\sum_{e'} C_{e', ty}) \right],
\end{equation}
where $C_{e, ty}$ is the confidence score returned by Probase and indicates how confident it believes that entity $e$ has a type $ty$.

After that, we calculate the similarity of two sibling entities using \Sg features as follows:
\begin{equation}\label{eq:skip-pattern-sim}
\small
sim_{sib}^{sk}(e_1, e_2 | SK) = \frac{\sum_{sk \in SK} \min(f_{e_1, sk}, f_{e_2, sk})}{\sum_{sk \in SK} \max(f_{e_1, sk}, f_{e_2, sk}) },
\end{equation}
where $SK$ denotes a selected set of ``discriminative'' \Sg features (see below for details). 
Similarly, we can calculate $sim_{sib}^{tp}(e_1, e_2)$ using all the type features.
Finally, we use the cosine similarity to compute the similarity between two entities based on their embedding features $sim_{sib}^{emb}(e_1, e_2)$.

To combine the above three similarities, we notice that a good pair of sibling entities should appear in similar contexts, share similar embeddings, and have similar types.
Therefore, we use a multiplicative measure to calculate the sibling similarity as follows:
\begin{equation}\label{eq:sim-sib}
\small
\begin{aligned}
sim_{sib}(e_1, e_2 | SK) &=  \sqrt{ (1+sim_{sib}^{sk}(e_1, e_2 | SK)) \cdot  sim_{sib}^{emb}(e_1, e_2)}  \\
	 &\cdot \sqrt{1+sim_{sib}^{tp}(e_1, e_2)}.
\end{aligned}
\end{equation}

\smallskip
\noindent \textbf{The Width Expansion Process.}
Given a seed entity set $S$ and a candidate entity list $V$, a straightforward idea to compute each candidate entity's average similarity with all entities in the seed set $S$ using all the features.
However, this approach can be problematic because (1) the feature space is huge (\ie, there are millions of possible \Sg features) and noisy, and (2) the candidate entity list $V$ is noisy in the sense that many entities in $V$ are completely irrelevant to $S$.
Therefore, we take a more conservative approach by first selecting a set of quality \Sg features and then scoring an entity only if it is associated with at least one quality \Sg feature.

Starting with the seed set $S$, we first score each \Sg feature based on its accumulated strength with entities in $S$ (\ie, $score(sk) = \sum_{e \in S} f_{e, sk}$), and then select top 200 \Sg features with maximum scores.
After that, we use sampling without replacement method to generate $10$ subsets of \Sg features $SK_{t}, t=1,2, \dots, 10$.
Each subset $SK_{t}$ has 120 \Sg features.
Given an $SK_{t}$, we will consider a candidate entity in $V$ only if it has association will at least one \Sg feature in $SK_{t}$.
The score of a considered entity is calculated as follows:

\begin{equation}\label{eq:entity-score}
\small
score(e|S, SK_{t}) = \frac{1}{|S|} \sum_{e' \in S} sim_{sib} (e, e' | SK_{t}).
\end{equation}

For each $SK_{t}$, we can obtain a rank list of candidate entities $L_{t}$ based on their scores.
We use $r_{t}^{i}$ to denote the rank of entity $e_{i}$ in $L_{t}$ and if $e_{i}$ does not appear in $L_{t}$, we set $r_{t}^{i} = \infty$.
Finally, we calculate the mean reciprocal rank ($mrr$) of each entity $e_{i}$ and add those entities with average rank above $r$ into the set $S$ as follows:
\small
\begin{equation}\label{eq:entity-score-mrr}
mrr(e_{i}) = \frac{1}{10}\sum_{t=1}^{10} \frac{1}{r_{t}^{i}}, \qquad S = S \cup \{e_{i} | mrr(e_{i}) > \frac{1}{r} \}.
\end{equation}
\normalsize
The key insight of above aggregation mechanism is that an irrelevant entity will not appear frequently in multiple $L_{t}$ at top positions and thus likely has a low $mrr$ score.
The same idea in proved effective in \cite{Shen2017SetExpanCS}. In this paper, we set $r=5$.

\subsubsection{Depth Expansion}\label{subsubsec:depth_expansion}
The width expansion algorithm requires an initial seed entity set to start with.
This requirement is satisfied for nodes in the initial seed taxonomy $\T^{0}$ as their children nodes can naturally form such a set.
However, for those newly-added nodes in taxonomy tree (\eg, the node \mquote{Canada} in Figure \ref{fig:overview}), they do not have any child node and thus we cannot directly apply the width expansion algorithm.
To address this problem, we use \emph{depth expansion} algorithm to acquire a target node's initial children by considering the relations between its sibling nodes and its niece/nephew nodes.
A concrete example is shown below.
\begin{example}[Depth Expansion]
Consider the node \mquote{Canada} in Figure \ref{fig:overview} as an example. 
This node is generated by the previous width expansion algorithm and thus does not have any child node.
We aim to find its initial children (\ie, \mquote{Ontario} and \mquote{Quebec}) by modeling the relation between the siblings of node \mquote{Canada} (\eg, \mquote{U.S.}) and its niece/nephew node (\eg, \mquote{California}, \mquote{Illinois}).
Similarly, given the target node \mquote{Mexico}, we want to find its initial children such as node \mquote{Sonora}.
\end{example}

Our depth expansion algorithm relies on term embeddings, which encode the term semantics in a fix-length dense vector.
We use $\v(t)$ to denote the embedding vector of term $t$.
As shown in \cite{Mikolov2013DistributedRO, Fu2014LearningSH, Luu2016LearningTE}, the offset of two terms' embeddings can represent the relationship between them, which leads to the following observation that $\v(``U.S") - \v(``California") \approx \v(``Canada") - \v(``Ontario")$.
Therefore, given a target parent node $e_{t}$, a set of reference edges $E = \{\langle e_{p}, e_{c} \rangle\}$ where $e_{p}$ is the parent node of $e_{c}$, we calculate the ``goodness'' of putting node $e_{x}$ under parent node $e_{t}$ as follows:
\small
\begin{equation}\label{eq:sim_par}
sim_{par}(\langle e_{t}, e_{x} \rangle) = cos\left(\v(e_{t})-\v(e_{x}), \frac{1}{|E|}\sum_{\langle e_{p}, e_{c} \rangle} \v(e_{p})-\v(e_{c})  \right),
\end{equation}
\normalsize
where $cos(\v(x), \v(y))$ denotes the cosine similarity between vector $\v(x)$ and $\v(y)$.
Finally, we score each candidate entity $e_{i}$ based on $sim_{par}(\langle e_{t}, e_{i} \rangle)$ and select top-3 entities with maximum score as the initial children nodes under node $e_{t}$.

The term embedding is learned from REPEL~\cite{Qu2018WeaklysupervisedRE}, a model for weakly-supervised Relation Extraction using Pattern-enhanced Embedding Learning.
It takes a few seed relation mentions (e.g. ``US-Illinois'' and ``US-California'') and outputs term embeddings as well as reliable relational phrases for target relation type(s). 
REPEL consists of a pattern module which learns a set of reliable textual patterns, and a distributional module, which learns a relation classifier on term representations for prediction. 
As both modules provide extra supervision for each other, the distributional module learns term embeddings supervised by more reliable patterns from the pattern module. 
By doing so, the learned term embeddings carry more useful information than those obtained from other embedding models like word2vec~\cite{Mikolov2013DistributedRO} and PTE~\cite{Tang2015PTEPT}, specifically for finding relation tuples of the target relation type(s).

\subsubsection{Conflict Resolution}\label{subsubsec:conflict_resolution}
Our hierarchical tree expansion algorithm \emph{iteratively} applies width expansion and depth expansion to grow the taxonomy tree to its full size.
As the supervision signal from the user-specified seed taxonomy $\T^{0}$ is very weak (\ie, only few nodes and edges are given), we need to make sure those nodes introduced in the first several iterations are of high quality and will not mislead the expansion process in later iterations to a wrong direction.
In this work, for each task-related term, we aim to find its single best position on our output task-guided taxonomy $\T$.
Therefore, when finding a term appears in multiple positions during our tree expansion process, we say a ``conflict'' happens and aim to resolve such conflict by finding the best position that term should reside in.

Given a set of conflicting nodes $\C$ which corresponds to different positions of a same entity, we apply the following three rules to select the best node out of this set.
First, if any node is in the seed taxonomy $\T^{0}$, we directly select this node and skip the following two steps.
Otherwise, for each pair of nodes in $\C$, we check whether one of them is the ancestor of the other and retain only the ancestor node.
After that, we calculate the ``confidence score'' of each remaining node $e \in C$ as follows:
\begin{equation}\label{eq:conf-score1}
\begin{aligned}
conf(e) &= \frac{1}{|sib(e)|} \sum_{e' \in sib(e)} sim_{sib}(e, e' | SK)  \\
	 &\cdot sim_{par}(\langle par(e), e \rangle),
\end{aligned}
\end{equation}
where $sib(e)$ denotes the set of all sibling nodes of $e$ and $par(e)$ represents its parent node.
The \Sg feature in $SK$ is selected based on its accumulated strength with entities in $sib(e)$.
This equation essentially captures a node's joint similarity with its siblings and its parent.
The node with highest confidence score will be selected.
Finally, for each node in $\C$ that is not selected, we will delete the whole subtree rooted by it, cut all the sibling nodes added after it, and put it in its parent node's ``children backlist".
A concrete example is shown below.
\begin{example}[Conflict Resolution]
In Figure \ref{fig:overview}, we can see there are two \mquote{Texas} nodes, one under \mquote{U.S.} and the other under \mquote{Mexico}.
As none of them is from initial ``seed'' taxonomy and they do not hold an ancestor-descendant relationship, we need to calculate each node's confidence score based on Eq.\ (\ref{eq:conf-score1}).
Since \mquote{Texas} has a stronger relation with other states in U.S., comparing with those in Mexico, we will select the \mquote{Texas} node under \mquote{U.S.}.
Then, for the other node under \mquote{Mexico}, we will delete it and cut \mquote{Coahuila}, a sibling node added after \mquote{Texas}.
Finally, we let the node \mquote{Mexico} to remember that \mquote{Texas} is not one of its children, which prevents the \mquote{Texas} node being added back later.
Notice that although the \mquote{Coahuila} node is cut here, it may be added back in a later iteration by our tree expansion algorithm.
\end{example}

\begin{algorithm}[!t]
  \caption{Hierarchical Tree Expansion.}
  \label{alg:hierarchical_tree_expansion}
  \KwIn{
    A seed taxonomy $\T^{0}$; a candidate term list $V$; maximum expansion iteration \emph{max\_iter}.
  }
  \KwOut{A task-guided taxonomy $\T$.}
  $\T \gets \T^{0}$\;
  \For{iter from 1 to max\_iter} {
  	$q \gets queue([\T.rootNode])$\;
	\While{q is not empty} {
		$e_{t} \gets q.pop()$\;
		$\boxdot$ \emph{\textbf{Depth Expansion}}\;
		\If{$e_{t}.children$ is empty} {
		 	$S \gets \depthExpan(e_{t})$\;
			$e_{t}.children \gets S$\;
			$q.push(S)$\;
		 }
		$\boxdot$ \emph{\textbf{Width Expansion}}\;
		$C_{new} \gets \widthExpan(e_{t}.children)$\;
		$e_{t}.children = e_{t}.children \oplus C_{new}$\;
		$q.push(C_{new})$\;
	}
	$\boxdot$ \emph{\textbf{Conflict Resolution}}\;
	Identify conflicting nodes in $\T$ and resolve the conflicts\;
  }
  Return $\T$\;
\end{algorithm}

\noindent\textbf{Summary.}
Algorithm \ref{alg:hierarchical_tree_expansion} shows the whole process of hierarchical tree expansion.
It iteratively expands the children of every node on a currently expanded taxonomy tree, starting from the root of this tree.
Whenever a target node $e_{t}$ with no children is found, it first applies depth expansion to obtain the initial children nodes $S$ and then uses width expansion to acquire more children nodes $C_{new}$.
At the end of each iteration, it resolves all the conflicting nodes.
The iterative process terminates after expanding the tree \emph{max\_iter} times and the final expanded taxonomy tree $\T$ will be returned.

\subsection{Taxonomy Global Optimization}\label{subsec:taxonomy_structure_opt}

In Algorithm \ref{alg:hierarchical_tree_expansion}, a node will be selected and attached onto the taxonomy based on its ``local'' similarities with other sibling nodes and its parent node.
While modeling only the ``local'' similarity can simplify the tree expansion process, we find the resulting taxonomy may not be the best from a ``global" point of view.
For example, when expanding the France regions, we find that the entity ``Molise'', an Italy region, will be mistakenly added under the ``France'' node, likely because it shares many similar contexts with some other regions of France.
However, when we take a global view of the taxonomy and ask the following question---\emph{which country is Molise located in?}, we can easily put ``Molise'' under ``Italy'' as it shares more similarities with those in Italy than in France.

Motivated by the above example, we propose a \emph{taxonomy global optimization module} in \HiExpan.
The key idea is to adjust each two contiguous levels of the taxonomy tree and to find the best ``parent'' node at the upper level for each ``child'' node at the lower level.
In Figure \ref{fig:overview}, for example, the upper level consists of all the countries while the lower level contains each country' first-level administrative divisions.
Intuitively, our taxonomy global optimization makes the following two hypotheses: (1) entities that have the same parent are similar to each other and form a coherent set, and (2) each entity is more similar to its correct parent compared with other siblings of its correct parent.

Formally, suppose there are $m$ ``parent'' nodes at the upper level and $n$ ``child'' nodes at the lower level, we use $\mathbf{W} \in \mathbb{R}^{n \times n}$ to model the entity-entity sibling similarity and use $\mathbf{Y^{c}} \in \mathbb{R}^{n \times p}$ to capture the two entities's parenthood similarity.
We let $\mathbf{W_{ij}} = sim_{sib}(e_{i}, e_{j})$ if $i \neq j$, otherwise we set $\mathbf{W_{ii}} = 0$.
We set $\mathbf{Y^{c}_{ij}} = sim_{par}(\langle e_{j}, e_{i}  \rangle)$.
Furthermore, we define another $n \times p$ matrix $\mathbf{Y}^{s}$ with $\mathbf{Y^{s}_{ij}} = 1$ if a child node $e_{i}$ is under parent node $e_{j}$ and $\mathbf{Y^{s}_{ij}} = 0$ otherwise.
This matrix captures the current parent assignment of each child node.
We use $\mathbf{F} \in \mathbb{R}^{n \times p}$ to represent the child nodes' parent assignment we intend to learn.
Given a $\mathbf{F}$, we can assign each ``child'' node $e_{i}$ to a ``parent'' node $e_{j} = \argmax_{j} \mathbf{F_{ij}}$.
Finally, we propose the following optimization problem to reflect the previous two hypotheses:
\begin{equation}\label{eq:opt_problem}
\small
\min_{\mathbf{F}} \sum_{i,j}^{n} \mathbf{W_{ij}} \left\Vert \frac{\mathbf{F_i}}{\sqrt{\mathbf{D_{ii}}}} - \frac{\mathbf{F_j}}{\sqrt{\mathbf{D_{jj}}}}  \right\Vert^{2}_{2}
+ \mu_{1} \sum_{i=1}^{n} \left\Vert \mathbf{F_{i}} - \frac{\mathbf{Y_{i}^{c}}}{ \| \mathbf{Y_{i}^{c}} \|_{1}   } \right\Vert^{2}_{2}
+ \mu_{2} \sum_{i=1}^{n} \left\Vert \mathbf{F_{i}} - \mathbf{Y_{i}^{s}} \right\Vert^{2}_{2},
\end{equation}
\normalsize
where $\mathbf{D_{ii}}$ is the sum of $i$-th row of $\mathbf{W}$, and $\mu_{1}, \mu_{2}$ are two nonnegative model hyper-parameters.
The first term in Eq. (\ref{eq:opt_problem}) corresponds to our first hypothesis and models two entities' sibling similarity.
Namely, if two entities are similar to each other (\ie, large $\mathbf{W_{ij}}$), they should have similar parent node assignments.
The second term in Eq. (\ref{eq:opt_problem}) follows our second hypothesis to model the parenthood similarity.
Finally, the last term in Eq. (\ref{eq:opt_problem}) serves as the smoothness constraints and captures the taxonomy structure information before the global adjustment.

To solve the above optimization problem, we take the derivative of its objective function with respect to $\mathbf{F}$ and can obtain the following closed form solution:
\begin{equation}\label{eq:conf-score}
\small
\begin{aligned}
\mathbf{F^{*}} &=  (\mathbf{I} - \alpha S  )^{-1} \cdot (\beta_{1} \mathbf{Y^{c}} + \beta_{2} \mathbf{Y^{s}}),  \\
\mathbf{S}	 &= \mathbf{D}^{-1/2}\mathbf{W}\mathbf{D}^{-1/2},
\end{aligned}
\end{equation}
where $\alpha_{1} = \frac{1}{1+\mu_{1}+\mu_{2}}$, $\beta_{1} = \frac{\mu_{1}}{1+\mu_{1}+\mu_{2}}$ and $\beta_{2} = \frac{\mu_{2}}{1+\mu_{1}+\mu_{2}}$. The calculation procedure is similar to the one in \cite{Zhou2003LearningWL}.

%% file: 5-experiments.tex
\section{Experiments}\label{sec:exp}

\subsection{Experimental Setup}

\subsubsection{Datasets}
We use three corpora from different domains to evaluate the performance of \HiExpan: (1) \textbf{DBLP} contains about 156 thousand paper abstracts in computer science field; (2) \textbf{Wiki} is a subset of English Wikipedia pages used in \cite{Ling2012FineGrainedER,Shen2017SetExpanCS}; (3) \textbf{PubMed-CVD} contains a collection of 463 thousand research paper abstracts regarding cardiovascular diseases retrieved from the PubMed\footnote{\url{https://www.ncbi.nlm.nih.gov/pubmed}.}. Table~\ref{tab:datasets} lists the details of these datasets used in our experiment. All datasets are available for download at: \url{http://bit.ly/2Jbilte}.

\begin{table}[!t]
    \caption{Datasets statistics.}\label{tab:datasets}
    \vspace{-0.1cm}
    \centering
    \begin{tabular}{cccc}
        \toprule
        \textbf{Dataset}    & \textbf{File Size} &  \textbf{\# of Sentences} & \textbf{\# of Entities} \\
        \midrule
        Wiki       &1.02GB      &1.50M      &41.2K                   \\
	DBLP       &520MB      &1.10M      &17.1K                    \\
	PubMed-CVD &1.60GB       &4.48M      &36.1K                  \\
	\bottomrule
    \end{tabular}
    \vspace{-0.3cm}
\end{table}

\subsubsection{Compared Methods}
To the best of our knowledge, we are the first to study the problem of task-guided taxonomy construction with user guidance, and thus there is no suitable baseline to compare with directly. 
Therefore, here we evaluate the effectiveness of \HiExpan by comparing it with a heuristic set-expansion based method and its own variations as follows:
\begin{itemize}[leftmargin=*]
    \item \HSetExpan is a baseline method which iteratively applies \SetExpan algorithm \cite{Shen2017SetExpanCS} at each level of taxonomy. For each lower level node, this method finds its best parent node to attach according to the children-parent similarity measure defined in Eq. (\ref{eq:sim_par}).
    \item \NoREPEL is a variation of \HiExpan without the REPEL \cite{Qu2018WeaklysupervisedRE} module which jointly leverages pattern-based and distributional methods for embedding learning. Instead, we use the SkipGram model~\cite{Mikolov2013DistributedRO} for learning term embeddings.
    \item \NoGTO is a variation of \HiExpan without the taxonomy global optimization module. It directly outputs the taxonomy generated by hierarchical tree expansion algorithm.
    \item \HiExpan is the full version of our proposed framework, with both REPEL embedding learning module and taxonomy global optimization module enabled. 
\end{itemize}

\subsubsection{Parameter Setting}
We use the above methods to generate three taxonomies, one for each corpus. 
When extracting the key term list using \AutoPhrase \cite{Shang2018AutomatedPM}, we treat phrases that occur over 15 times in the corpus to be frequent. 
The embedding dimension is set to 100 in both REPEL \cite{Qu2018WeaklysupervisedRE} and SkipGram model \cite{Mikolov2013DistributedRO}. 
The maximum expansion iteration number \textit{max\_iter} is set to 5 for all above methods. 
Finally, we set the two hyper-parameters used in taxonomy global optimization module as $\mu_{1} = 0.1$ and $\mu_{2} = 0.01$.

\subsection{Qualitative Results}
\begin{figure*}[!th]
  \centering {\
    \subfigure[Parts of the taxonomy generated by \HiExpan on the Wiki dataset.] {
      \label{fig:tax-wiki}
      \includegraphics[width=0.8\textwidth]{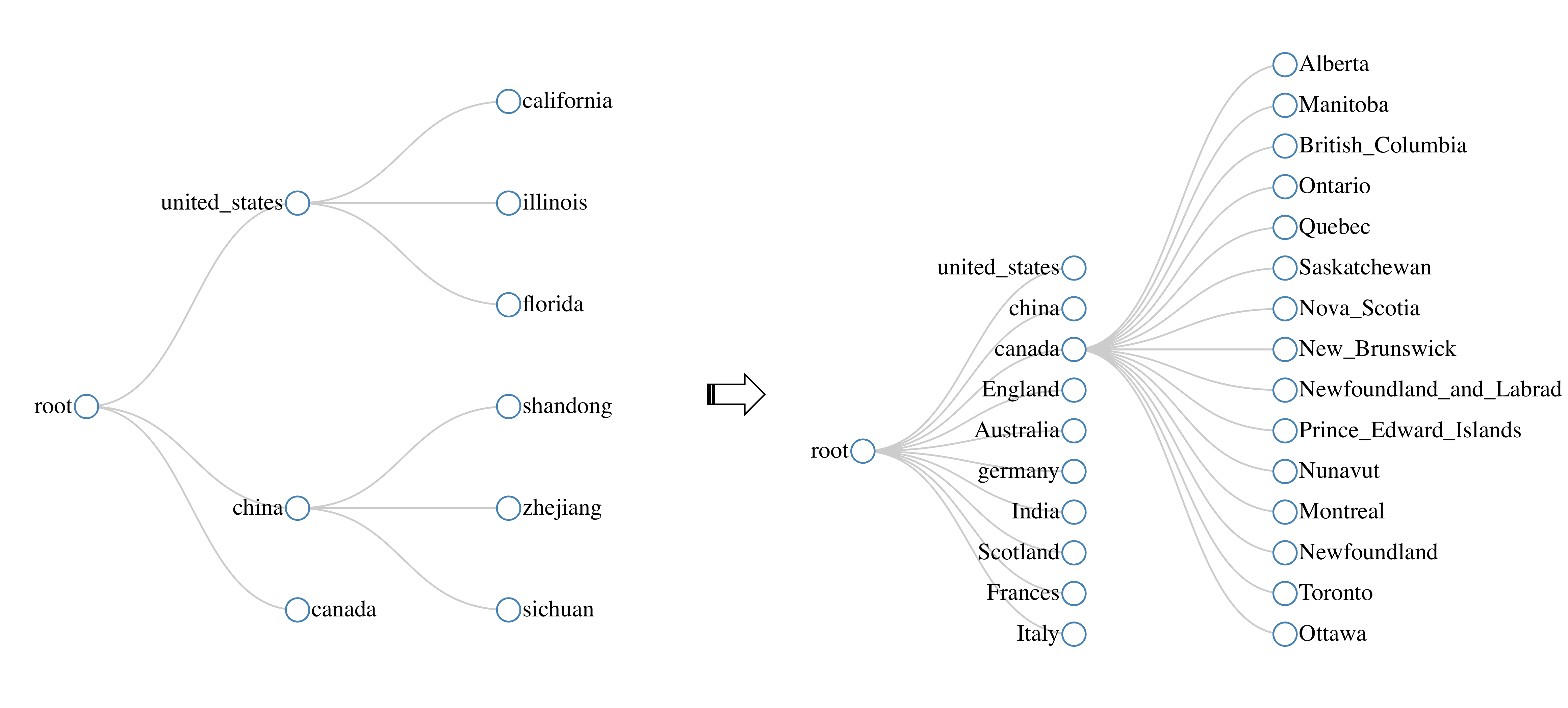}
    }
    \vspace{-1ex}
    \hfil
    \subfigure[Parts of the taxonomy generated by \HiExpan on the DBLP dataset.] {
      \label{fig:tax-dblp}
      \includegraphics[width=0.9\textwidth]{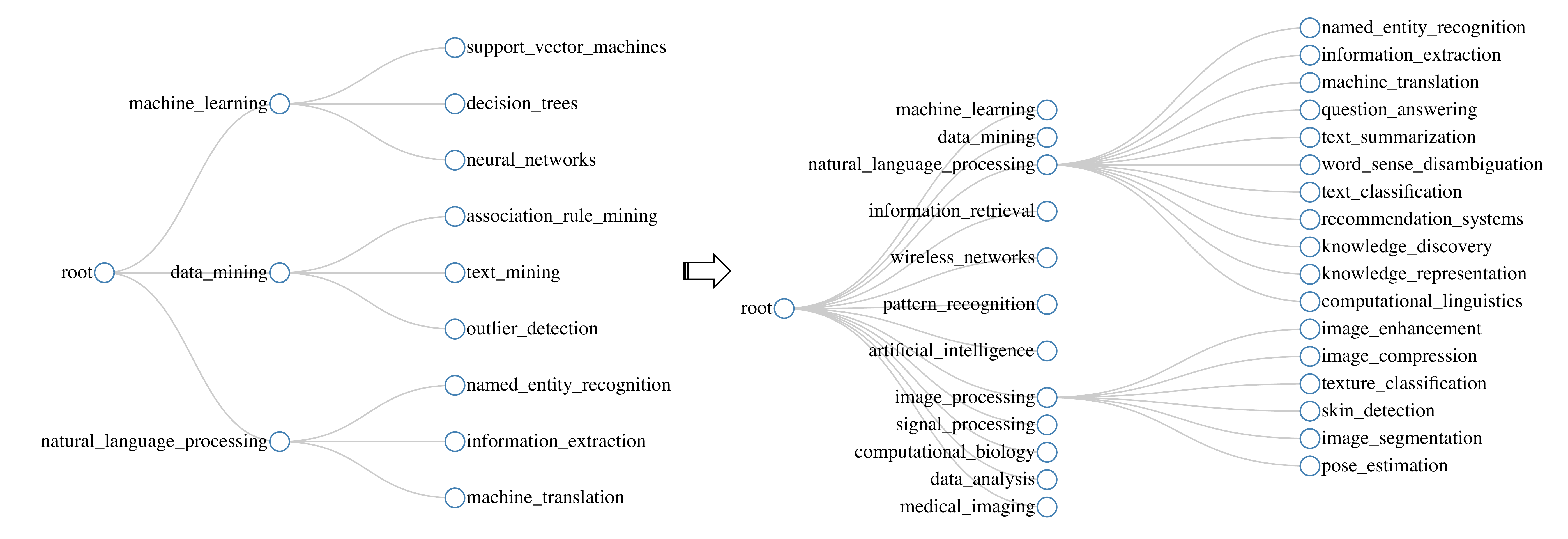}
    }
    \vspace{-1ex}
    \hfil
    \subfigure[Parts of the taxonomy generated by \HiExpan on the PubMed-CVD dataset.] {
      \label{fig:tax-cvd}
      \includegraphics[width=0.9\textwidth]{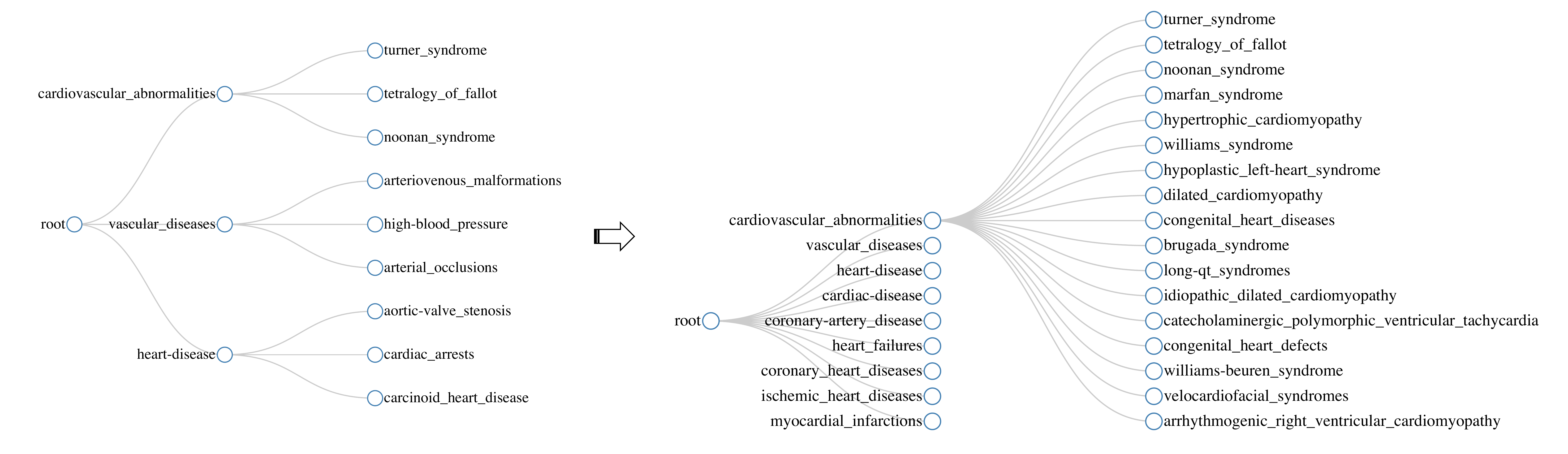}
    }
  }
  \vspace{-1ex}
  \caption{Qualitative results: we show the taxonomy trees generated by \HiExpan across three different corpora.}
  \label{fig:three_taxonomies}
\end{figure*}

In this subsection, we show the taxonomy trees generated by \HiExpan across three text corpora with different user-guidances. 
Those seed taxonomies are shown in the left part of Figure \ref{fig:three_taxonomies}. 
\begin{itemize}[leftmargin=*]
    \item As shown in Figure \ref{fig:tax-wiki}, the ``seed'' taxonomy containing three countries and six states/provinces. At the first level, we have ``United States'', ``China'' as well as ``Canada''. 
    Under the node "United States", we are given ``California'', ``Illinois'', as well as ``Florida'' as initial seeds. We do the same for ``Shandong'', ``Zhejiang'' and ``Sichuan'' under node ``China''. 
    Our goal is to output a taxonomy which covers all countries and state/provinces mentioned in the corpus and connects them based the ``country-state/province'' relation. 
    On the right part of Figure \ref{fig:tax-wiki}, we show a fragment of the taxonomy generated by \HiExpan which contains the expanded countries and Canadian provinces. 
    \HiExpan first uses the depth expansion algorithm to find initial children under ``Canada'' (\ie, ``Alberta'' and ``Manitoba'') and then, starting from the set \{``Alberta'', ``Manitoba''\}, it applies the width expansion algorithm to obtain more Canadian provinces. 
    These steps are repeated and finally \HiExpan is able to find countries like ``England'', ``Australia'', ``Germany'' in the first-level of taxonomy and to discover states/provinces of each country. 
        
    \item Figure \ref{fig:tax-dblp} shows parts of the taxonomy generated by \HiExpan on the DBLP dataset. 
    Given the initial seed taxonomy (the left part of Figure \ref{fig:tax-dblp}), \HiExpan automatically discovers many computer science subareas such as ``information retrieval'', ``wireless networks'' and ``image processing''. 
    We can also zoom in to look at the taxonomy at a more granular level. 
    Taking the node ``natural language processing'' as an example, \HiExpan successfully finds major subtopics in natural language processing such as ``question answering'', ``text summarization'', and ``word sense disambiguation''. 
    \HiExpan can also find subtopics under image processing even without any initial seeds entities. 
    As shown on the right part of Figure \ref{fig:tax-dblp}, we have obtained high-quality subtopics of ``image processing'' such as ``image enhancement'', ``image compression'', ``skin detection'', and etc. 

    \item In Figure \ref{fig:tax-cvd}, we let \HiExpan to run on the PubMed-CVD data and show parts of the resulting taxonomy. 
    We feed the model with 3 seeds at the top level, namely ``cardiovascular abnormalities'', ``vascular diseases'' and ``heart disease'' along with 3 seeds under each top-level node. At the top level, \HiExpan generates labels such as ``coronary artery diseases'', ``heart failures'', ``heart diseases'', and ``cardiac diseases''. 
    Here, we notice that many labels, \eg, ``heart disease'' and ``cardiac disease'' are actually synonyms. 
    These synonyms are put at the same level in the taxonomy generated by \HiExpan since they share same semantics and appear in similar contexts. 
    We leave synonyms discovery and resolution as an important future work.
\end{itemize}


\begin{table*}[t]
    \caption{\NoGTO shows the parent of an entity before applying taxonomy structure optimization. \HiExpan shows the parent node of this entity after optimizing the taxonomy structure.}\label{tbl:case_study}
    \vspace{-0.1cm}
    \centering
    \small
    \begin{tabular}{cccc}
        \toprule
        \textbf{Dataset} & \textbf{Entity} & \textbf{\textsf{NoGTO}} & \textbf{\textsf{HiExpan}} \\
        \midrule
        \multirow{5}{*}{\textsf{Wiki}} & London & Australia & England \\ 
        & Chiba & China & Japan \\
	& Molise & Frances & Italy \\
	& New\_South\_Wales & England & Australia \\
	& Shropshire & Scotland &  England \\
        \midrule
        \multirow{5}{*}{\textsf{DBLP}} & unsupervised\_learning & data\_mining & machine\_learning \\
	& social\_network\_analysis & natural\_language\_processing & data\_mining \\
	& multi-label\_classification & information\_retrieval &  machine\_learning \\
	& pseudo-relevance\_feedback & computational\_biology &  information\_retrieval  \\
	& function\_approximate & data\_analysis & machine\_learning \\
        \bottomrule
    \end{tabular}
    \label{fig:gto_effect}
\end{table*}

Table \ref{fig:gto_effect} shows the effect of taxonomy global optimization module in \HiExpan. 
From the experiment on the Wiki dataset, we observe that `the node `London'' was originally attached to ``Australia'', but after applying the taxonomy global optimization module, this node is correctly moved under ``England''. 
Similarly, in the DBLP dataset, the term ``unsupervised learning'' was initially located under ``data mining'' but later being moved under the parent node ``machine learning''. 
This demonstrates the effectiveness of our taxonomy global optimization module.

\subsection{Quantitative Results}
In this subsection, we quantitatively evaluate the quality of the taxonomies constructed by different methods. 

\begin{table*}[!t]
\centering
\caption{Quantitative results: we show the quantitative results of the taxonomies constructed by \HSetExpan, \NoREPEL, \NoGTO, and \HiExpan. 
$P_{a}$, $R_{a}$, $F1_a$ denote the ancestor-Precision, ancestor-Recall, and ancestor-F1-score, respectively. Similarly, we denote edge-based metrics as $P_e$, $R_e$, and $F1_e$, respectively.
}
\vspace{-0.3cm} 
\scalebox{0.83}{
        \begin{tabular}{ccccccc|cccccc|cccccc}
            \toprule
            \multirow{3}{*}{\textbf{Method}} & \multicolumn{6}{c}{\textbf{Wiki}} & \multicolumn{6}{c}{\textbf{DBLP}} & \multicolumn{6}{c}{\textbf{PubMed-CVD}}  \\
            \cmidrule{2-19}
                           & $P_{a}$ & $R_{a}$ & $F1_{a}$ & $P_{e}$ & $R_{e}$ & $F1_{e}$       & $P_{a}$ & $R_{a}$ & $F1_{a}$ & $P_{e}$ & $R_{e}$ & $F1_{e}$       & $P_{a}$ & $R_{a}$ & $F1_{a}$ & $P_{e}$ & $R_{e}$ & $F1_{e}$  \\ 
                           \midrule
\HSetExpan        &    0.740      &    0.444      &   0.555        &    0.759       &     0.471     &    0.581            &   0.743        &   \textbf{0.448}        & \textbf{0.559}            &    0.739       &   0.448        &    0.558               &  0.524        & 0.438        &   0.477        &   0.513       &   0.459        &  0.484         \\ 
\midrule
\NoREPEL         &   0.696         &   0.596      &   0.642        &   0.697        &  0.576        &  0.631              &  0.722         &  0.384         & 0.502            &   0.705        &  \textbf{0.464}         &   0.560                &   0.583       &    \textbf{0.473}        &   0.522        &   0.593        &    \textbf{0.541}       &  0.566       \\ 
\NoGTO             &    0.827        &    0.708       &   0.763          &      0.810     &   0.671       &   0.734             &   0.821      &    0.366       & 0.506           &  0.779         &   0.433        &    0.556     &    0.729       &     0.443       &   0.551          &   0.735      &   0.506         &  0.599    \\ 
\HiExpan            &    \textbf{0.847}        &    \textbf{0.725}       &     \textbf{0.781}        &    \textbf{0.848}       &       \textbf{0.702}   &       \textbf{0.768}         &  \textbf{0.843}      & 0.376          & 0.520           &   \textbf{0.829}        &  0.460         &    \textbf{0.592}     &   \textbf{0.733}        &   0.446         &   \textbf{0.555}      &   \textbf{0.744}       &   0.512        &  \textbf{0.606}             \\ 
            \bottomrule
        \end{tabular}
}
\label{tab:quantitative_results}
\vspace{0.3cm}
\end{table*}

\subsubsection{Evaluation Metrics}
Evaluating the quality of an entire taxonomy is challenging due to the existence of multiple aspects that should be considered and the difficulty of obtaining gold standard~\cite{Wang2017ASS}. 
Following~\cite{task17semeval2015, bordea2016semeval, Mao2018EndtoEndRL}, we use $\ancestorF1$ and $\edgeF1$ for taxonomy evaluation in this study. 

\textbf{Ancestor-F1} measures correctly predicted ancestral relations. It enumerates all the pairs on the predicted taxonomy and compares these pairs with those in the gold standard taxonomy.
\begin{equation*}
\begin{aligned}
P_{a} &= \frac{| \text{is-ancestor}_{\text{pred}} \cap \text{is-ancestor}_{\text{gold}} |} {| \text{is-ancestor}_{\text{pred}} |} ,\\
R_{a} &= \frac{| \text{is-ancestor}_{\text{pred}} \cap \text{is-ancestor}_{\text{gold}} |} {| \text{is-ancestor}_{\text{gold}} |} ,\\
F1_{a} &= \frac{2P_{a}*R_{a}}{P_{a}+R_{a}} ,
\end{aligned}
\end{equation*}
where $P_{a}$, $R_{a}$, $F1_a$ denote the ancestor precision, ancestor recall, and ancestor F1-score, respectively.

\textbf{Edge-F1} compares edges predicted by different taxonomy construction methods with edges in the gold standard taxonomy.
Similarly, we denote edge-based metrics as $P_e$, $R_e$, and $F1_e$, respectively.

To construct the gold standard, we extract all the parent-child edges in taxonomies generated by different methods in table \ref{tab:quantitative_results}. Then we pool all the edges together and ask five people, including the second and third author of this paper as well as three volunteers, to judge these pairs independently. We show them seed parent-child pairs as well as the generated parent-child pairs, and ask them to evaluate whether the generated parent-child pairs have the same relation as the given seed parent-child pairs. After collecting these answers from the annotators, we simply use majority voting to label the pairs. We then use these annotated data as the gold standard. The labeled dataset is available at: \url{http://bit.ly/2Jbilte}.

\subsubsection{Evaluation Results}

Table \ref{tab:quantitative_results} shows both the ancestor-based and edge-based precision/recalls as well as F1-scores of different methods. 
We can see that \HiExpan achieves the best overall performance, and outperforms other methods, especially in terms of the precision.
Comparing the performance of \HiExpan, \NoREPEL, and \NoGTO, we see that both the REPEL and the taxonomy global optimization modules play important roles in improving the quality of the generated taxonomy. 
Specifically, REPEL learns more discriminative representations by iteratively letting the distributional module and pattern module mutually enhance each other, and the taxonomy global optimization module leverages the global information from the entire taxonomy tree structure. 
In addition, \HiExpan resolves the ``conflicts'' at the end of each tree expansion iteration by cutting many nodes on a currently expanded taxonomy. 
This leads \HiExpan to generate a smaller tree comparing with the one generated by \HSetExpan, given that both methods running the same number of iterations. 
However, we can see that \HiExpan still beats \HSetExpan on Wiki dataset and PubMed-CVD dataset, in terms of the recall. 
This further demonstrates the effectiveness of our \HiExpan framework.

%% file: 6-conclusion.tex

\section{Conclusions and Future Work}\label{sec:con}

In this paper, we introduce a new research problem \emph{task-guided taxonomy construction} and propose a novel expansion-based framework \HiExpan for solving it.
\HiExpan views all children under a taxonomy node as a coherent set and builds the taxonomy by recursively expanding these sets.
Furthermore, \HiExpan incorporates a weakly-supervised relation extraction module to infer parent-child relation and adjusts the taxonomy tree by optimizing its global structure.
Experimental results on three public datasets corroborate the effectiveness of \HiExpan.

As a first-punch solution for constructing a task-guided taxonomy, \HiExpan can be improved in many ways.
First, we find in the experiments that \HiExpan places synonyms at the same level of taxonomy since they share same semantic meanings and appear in similar contexts.
These synonyms will make generated taxonomy less informative, with reduced overall quality.
It is an interesting direction to extend \HiExpan to automatically discover and resolve those synonyms.
Further, as an expansion-based framework, \HiExpan may facilitate interactive user guidance in taxonomy construction, which is another interesting task in the future. 

%% file: 7-ack.tex

\section*{Acknowledgements}\label{sec:ack}
This research is sponsored in part by U.S. Army Research Lab. under Cooperative Agreement No. W911NF-09-2-0053 (NSCTA), DARPA under Agreement No. W911NF-17-C-0099, National Science Foundation IIS 16-18481, IIS 17-04532, and IIS-17-41317, DTRA HDTRA11810026, and grant 1U54GM114838 awarded by NIGMS through funds provided by the trans-NIH Big Data to Knowledge (BD2K) initiative (www.bd2k.nih.gov). 
We thank Xinwei He, Yunyi Zhang, and Luyu Gao for helping label the datasets and providing valuable comments and discussions. 
Also, we would like to thank anonymous reviewers for valuable feedback.